\documentclass[wcp]{jmlr}

\usepackage{amsmath}
\usepackage{mathtools}
\usepackage{bbold}
\usepackage{mathrsfs}
\usepackage{bm}
\usepackage{dsfont}\usepackage{capt-of}
\usepackage{booktabs}
\usepackage{varwidth}
\usepackage{capt-of}

\DeclareMathAlphabet{\pazocal}{OMS}{zplm}{m}{n}

\newcommand{\re}{\mathbb{R}}
\newcommand{\ex}{\mathbb{E}}

\newcommand{\matX}{\mathbf{X}}
\newcommand{\matW}{\mathbf{W}}

\newcommand{\matZ}{\mathbf{Z}}
\newcommand{\matA}{\mathbf{A}}
\newcommand{\matI}{\mathbf{I}}
\newcommand{\vecz}{\mathbf{z}}
\newcommand{\vect}{\mathbf{t}}
\newcommand{\vecx}{\mathbf{x}}

\usepackage{booktabs}

\makeatletter
\let\Ginclude@graphics\@org@Ginclude@graphics 
\makeatother

\usepackage{caption, amsthm} 
\newtheorem{prop}{Proposition}

\usepackage{longtable}

\usepackage{booktabs}

\makeatother

\jmlrvolume{157}
\jmlryear{2021}
\jmlrworkshop{ACML 2021}

\title[DAGSurv]{DAGSurv: Directed Acyclic Graph Based Survival Analysis Using Deep Neural Networks}

 \author{\Name{Ansh Kumar Sharma*} \Email{ansh18130@iiitd.ac.in}\\
    \Name{Rahul Kukreja*} \Email{rahul18254@iiitd.ac.in}     
     \thanks{* indicates equal contribution}\\
    \Name{Ranjitha Prasad} \Email{ranjitha@iiitd.ac.in}\\
  \addr ECE dept., IIIT Delhi\\
  \Name{Shilpa Rao} \Email{shilpa@iiitg.ac.in}\\
  \addr ECE dept., IIIT Guwahati}

\editors{Vineeth N Balasubramanian and Ivor Tsang}

\begin{document}

\maketitle

\begin{abstract}
Causal structures for observational survival data provide crucial information regarding the relationships between covariates and time-to-event. We derive motivation from the information theoretic source coding argument, and show that incorporating the knowledge of the directed acyclic graph (DAG) can be beneficial if suitable source encoders are employed. As a possible source encoder in this context, we derive a variational inference based conditional variational autoencoder for causal structured survival prediction, which we refer to as \texttt{DAGSurv}. We illustrate the performance of \texttt{DAGSurv} on low and high-dimensional synthetic datasets, and real-world datasets such as METABRIC and GBSG. We demonstrate that the proposed method outperforms other survival analysis baselines such as \texttt{Cox} Proportional Hazards, \texttt{DeepSurv} and \texttt{Deephit}, which are oblivious to the underlying causal relationship between data entities.
\end{abstract}

\section{Introduction}
Modern data analysis and processing involve vast amounts of data, where the structure carries critical information about the interrelationships between the entities. This structure is often captured via a \emph{graph}, where an unweighted/weighted edge provides a flexible way of representing the relationship between the nodes. Several signal processing and machine learning algorithms in the past decade have analyzed graphical data \citep{marques2020graph}. In the context of machine learning, ignoring these relationships between covariates in the data may lead to biased and erroneous predictions. Hence, it is crucial to incorporate the knowledge of graph topology into learning algorithms. 

\begin{figure}[h]
    \centering
    \includegraphics[width=\textwidth]{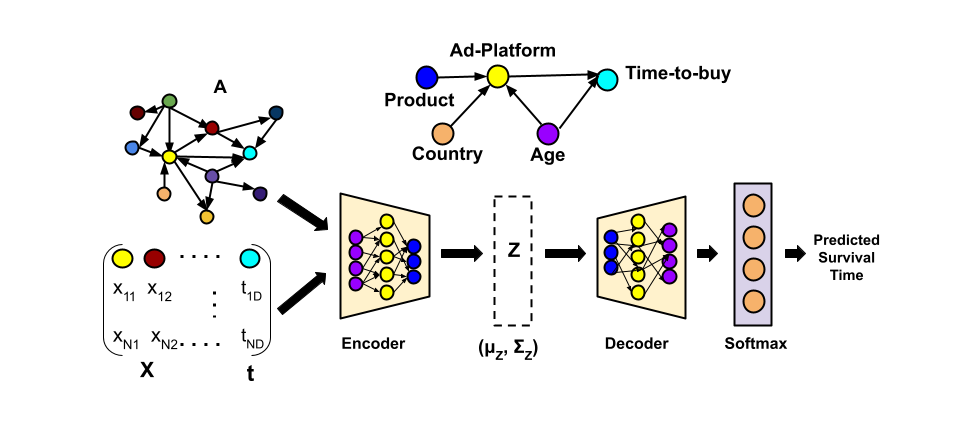}
    \caption{DAGSurv framework: The input of conditional VAE consists of the dataset $\mathcal{D}$ (defined in the sequel) and the adjacency matrix $\matA$. The latent variable that encodes $\mathcal{D}$ and $\matA$ is given by $\matZ$. Unlike conventional VAE, the output of CVAE based decoder is the conditional distribution $p(\vect|\matX,\matZ)$, and we apply a softmax layer to obtain the predicted survival time. We also illustrate the example graph from the advertising use-case. }
    \label{fig:DeepDAGSurvModel}
\end{figure}

Directed acyclic graphs (DAG) allows statistical modeling of  covariates by enforcing a topological ordering of these entities. DAGs are useful in answering  what-if questions such as “\emph{What} would be the system behavior \emph{if} a variable is set to a value A instead of B?”, with a focus on taking actions that induce a controlled change in systems. For instance, while placing an advertisement on online platforms, the relevant \emph{what-if} question is associated with the platform used for ad-placement, and the outcome is time-to-buy. Obtaining the cause-effect relationship between the platform and the outcome only would lead to erroneous predictions since user covariates such as age, geography, online purchase behavior, economic strata etc., also impact a purchase, albeit indirectly \citep{Camta}, as depicted in Fig.~\ref{fig:DeepDAGSurvModel}. Modeling such data as a graphical model allows us to encode the graph structure using conditional independence relationship among random variables that are represented by the vertices, as depicted in Fig.~\ref{fig:DeepDAGSurvModel}. In this work, we assume that the joint distribution of the covariates factorizes as dictated by the adjacency matrix of a DAG whose vertices are features of the dataset. 

Survival analysis (SA) is a well-known statistical technique for the study of temporal events, where  time-to-an-event data is modeled using a probabilistic function of fully or partially observed covariates. An impediment in modeling time-to-event data is the presence of \emph{censored} observations, i.e., instances whose event of interest is not observed (and hence, time-to-event information is missing). Neglecting censored data introduces bias in the inference process, and hence, analyzing such data necessitates significantly different statistical and machine learning techniques \citep{katzman2018deepsurv,lee2018deephit}. Moreover, such maximum likelihood techniques for survival analysis do not enforce any relationship between the features, and the model learns the interactions between the features and the time-to-event outcomes, i.e., any feature interaction is outcome based. In our work, we provide the DAG as an input, with the features as the nodes of the DAGs and their interactions is represented by the edges of the DAG.

\textbf{Contributions:} In this work, we integrate the cause-effect relationship between covariates and the time-to-event outcome by encoding the causal DAG structure into the analysis of temporal data. The contributions are as follows: 
\begin{itemize}
    \item Using information-theoretic source coding arguments, we show that by utilizing the knowledge of the adjacency matrix along with the input covariates leads to optimal encoding of the source distribution as compared to the case where covariates are assumed to be statistically independent. 
    \item Motivated by the source coding argument, we propose a conditional variational autoencoder (CVAE) based novel deep-learning architecture to incorporate the knowledge of the causal DAG for structured survival prediction, which we refer to as \texttt{\textbf{DAGSurv}}.
    \item We demonstrate the performance of the proposed \texttt{DAGSurv} framework using the time-dependent concordance index (CI) as a metric, on synthetic and real-world datasets such as Metabric and GBSG.
\end{itemize}
 Using experimental results, we demonstrate that incorporating the causal DAG in survival prediction is beneficial, not only for improving outcomes but also for validating the assumed causal dynamics of a system. In the case of real-world datasets, DAG is not readily available and hence, we use a pre-processing step where we estimate the graph from the given dataset, and use the estimated graph as an input to the proposed model. Simulation results confirm our hypothesis that incorporating the DAG into the machine learning model indeed leads to better representation of data which further leads to improved values of time-dependent CI, as compared to conventional SA techniques.

In the sequel, we describe the mathematical preliminaries of SA followed by the source coding argument for optimal source compression if the adjacency matrix is known. Subsequently, we define the proposed \texttt{DAGSurv} framework, and conclude with experimental results and discussions.
\section{Related Works}

It has been proven time and again that incorporating the knowledge of the graph structure into machine learning models reaps immense benefits. Graph convolutional networks (GCNs) are powerful tools that are used with undirected graphs for semi-supervised classification per instance in the dataset \citep{kipf2016semi}. In this work, we focus on exploiting the relationship between the covariates in a dataset, and hence, the GCN is not directly applicable. Knowledge graphs bring in the ability to establish relationships between entities in an efficient manner that is explainable and re-usable. However, these relationships are often semantic \citep{nickel2015review}, and may not be of statistical relevance. 

In cases where graphs provide statistical information, probabilistic graphical models framework play an important role \citep{koller2009probabilistic}. In probabilistic graphical models, nodes of a graph are considered as random variables, and the covariate and target information are considered as the realizations of these random variables. Evidently, the edge between the random variables translates the statistical relationships between random variables, and hence, the graph forms a joint distribution over the dataset. In scenarios where the underlying graph is known, deep neural networks have been used along with graphical models for prediction \citep{yoon2019inference}. In this work, we utilize the probabilistic graphical models based framework for graph-based survival prediction. 

In the context of survival analysis, Kaplan-Meier (KM) technique is a popular but naive, covariate-ignorant non-parametric technique for obtaining the empirical estimate of the survival function\citep{kaplan1958nonparametric}. An improvement to the KM technique is the Cox proportional hazards model \citep{cox2018analysis} (CPH) which incorporates the user covariates for inference. Several parametric methods that propose Weibull or log-normal distributions \cite{wang2019machine} and non-parametric methods using Gaussian processes have been proposed for survival analysis \citep{fernandez2016gaussian}. Modern techniques based on deep neural networks (DNNs) have been used for time-to-event analysis in \citep{faraggi1995neural} and \citep{katzman2018deepsurv}, where non-linear representations replace linear models for modeling the relationship between covariates and the risk. However, the limitation of these methods is the inherent assumption of constant hazard rate and the linearity of the log-hazard rate. In \citep{lee2018deephit}, authors propose a cumulative index curve (CIC) approach, which uses the marginal probabilities of an event, in the presence of multiple competing events. This technique does not assume constant hazard rate or any other assumptions about the model.

Probabilistic graphical models have been used in the context of survival analysis \citep{bandyopadhyay2015data} where graph based inference algorithms are proposed for survival prediction assuming constant hazard rate. In contrast, we propose a conditional VAE (CVAE) based graphical model approach for structured survival prediction, where we do not assume constant hazard rate. Our work is closely related to DAG-GNN \citep{YuCGYicml19}. Note that the proposed CVAE is inspired by certain design aspects of DAG-GNN, but it is substantially different in functionality, as compared to DAG-GNN \citep{YuCGYicml19}. In DAG-GNN, the VAE (and not CVAE) is designed to  learn the weighted adjacency matrix of the DAG and it is not specific to a machine learning task. In our work, we incorporate the adjacency matrix as a \emph{known} parameter, and subsequently obtain an assumption-free machine learning model for survival prediction. Although, survival analysis is the theme of this work, it will be evident from the analysis that our model can be adapted for classification and regression tasks as well. 

Several methods that incorporate graph-represented relations of features into predictions approaches using GCNs have been proposed in literature. However, these methods incorporate separate modules for graph embedding and regression, classification or survival analysis. For instance, in \citep{di2020ranking}, a graph is considered between patches of pathological images and the feature representation generated by GCN is considered for survival analysis. On the other hand, we embed the knowledge of the graph into the network, and specifically address the problem of survival analysis. Another closely related work is \citep{chen2019survival}, where authors propose an undirected graph based survival analysis by using a probabilistic graphical model with the exponential family distribution to describe the relationship between the covariates. In comparison, we specifically consider DAGs to model causal relationships, and do not assume specific probabilistic models among covariates.  

\section{DAG Based Survival Analysis: Preliminaries and Loss Function}

In this section, we describe the problem of DAG-based SA. First, we provide mathematical preliminaries of survival prediction and subsequently formulate the problem based on the source coding argument. We propose the CVAE framework as a possible source encoder that incorporates the knowledge of DAG for survival prediction. We develop the variational loss function, which is dual-purpose in the sense that it incorporates the causal DAG along with learning system parameters for survival prediction.

\subsection{Mathematical Preliminaries}

Time-to-event datasets such that the dataset $\mathcal{D} = \{(\mathbf{x}^{(n)}, t^{(n)} ,\delta^{(n)})\}_{n = 1}^N$ are usually characterized by three variables for the $n$-th instance where, $\mathbf{x}^{(n)} \in \mathbb{R}^{L}$, i.e, for $n$ instances, $\matX \in {\re}^{N\times L}$. Here, $L$ represents the number of covariates. We consider survival time $t^{(n)}$ as discrete, and the time horizon as finite so the $t \in \mathcal{T}$ where $\mathcal{T} = \{0,\hdots,M\}$ for a predefined maximum time horizon $M$. Further, $\vect \in {\re}^{N\times 1}$ represents the time at which the event has occurred and $\delta^{(n)} \in \{0,1\}$ is an indicator variable which specifies if the $n$-th instance is censored or not. Time-to-event models are characterized by the survival function given by
\begin{equation*}
    S(t|\vecx)= P(T>t|\vecx)= 1-F(t|\vecx),
\end{equation*}
which is defined as the fraction of the population that survives up to time $t$ \footnote{For better readability, we drop the superscript $n$ while discussing about generic concepts.}, where $F(t|\vecx)$ represents the cumulative distribution function of time-to-event, given user covariates $\vecx$. Another important statistic is the conditional hazard rate function $h(t|\vecx)$ which is defined as the instantaneous rate of occurrence of an event at time $t$ given covariates $\vecx$. It is known that the relationship between $h(t|\vecx)$ and $S(t|\vecx)$ from standard definitions is given by:
\begin{equation}
    h(t|\vecx) = \lim_{dt \rightarrow 0} \frac{P(t < T < t+dt|\vecx)}{P(T > t|\vecx)dt}= \frac{f(t|\vecx)}{S(t|\vecx)},   
\end{equation}
where $f(t|\vecx)$ is the conditional survival density function and $S(t|\vecx)$ is as described earlier. The \texttt{Cox}-PH model \cite{cox1972regression} is a semi-parametric, linear model where the conditional hazard function $h(t|\vecx)$ depends on time through the baseline hazard $h_0(t)$, and independent covariates $\vecx$ such that
\begin{equation}
    h(t|\vecx) = h_0(t)\exp(\vecx^T\bm{\gamma}).
\end{equation}
For a given dataset with $N$ observations as described earlier, \texttt{Cox}-PH estimates the regression coefficients, $\bm{\gamma} \in \mathbb{R}^L$, such that the partial likelihood is maximized  \citep{cox1972regression}. In \texttt{DeepSurv}, the authors propose a CPH based DNN, as the basis for a treatment recommender system. Further, \texttt{DeepHit} directly learns the joint distribution of survival times and events, effectively avoiding the PH  assumptions or those inherent to a particular form of the model. In these methods, the covariates are assumed to be independent, and there is no formal mechanism using which any dependence between covariates can be included. In \cite{chen2019survival}, an undirected graph is assumed between the covariates and an exponential distribution based probabilistic graphical model is incorporated into analysis. However, in contrast, we design a CVAE based framework for incorporating a DAG between the covariates for survival prediction. Note that the proposed technique does not require any modeling assumptions such as those in \cite{chen2019survival}, and hence, it is suitable for all datasets. 

\subsection{Problem Formulation}

In this work, we employ the the DAG, denoted as $\mathcal{G}(V, E)$, to describe the causal relationship between the features in the dataset $\mathcal{D}$. Each vertex  in $\mathcal{G}(V,E)$ represents a random variable with $V = \{1, \hdots , L+1\}$ consisting of the indices of these random variables, i.e., $X_l$ is a vertex if $l \in V$. Further, let $V \times V$ consist of all pairs of indices in $V$. A pair of random variables $\{X_l, X_m\}$ is called an edge of the graph $\mathcal{G}$ if $(l, m) \in E \subset V \times V$. The $L+1$ vertices includes the $L$ covariates in $\matX$, and the $L+1$-th vertex is the target variable given by the survival time $\vect$. Let $\matA \in \re^{\left(L+1 \right)\times \left(L+1 \right)}$ denote the weighted adjacency matrix of this DAG. 

\subsubsection{Motivation} 

In this work, the covariate matrix $\matX$ and the adjacency matrix $\matA$ are encoded into an efficient representation for  structured survival prediction. We view the problem of encoding $\matX$ and $\matA$ jointly as a problem of \emph{source encoding}. We invoke the basic principles of information theory which establishes the fundamental limit for the compression of information. For optimal source compression, the expected length of the source code must be greater than or equal to the entropy of the source \citep{cover1999elements}. First we note that the adjacency matrix governs the probabilistic relationship between the features, as given by the following proposition.  

\begin{prop}
The adjacency matrix $\matA$ of the directed acyclic graph (DAG) $\mathcal{G}(V,E)$ characterizes the joint distribution $p(\vect,\matX)$.
\label{prop1}
\end{prop}
\begin{proof}
See the supplementary material.
\end{proof}

In the next two propositions, we establish that the entropy of the source that emits symbols governed by $p(\vect,\matX|K_{\matA})$ with $\matA \neq 0$, is upper bounded by the entropy of a source that emits statistically independent source symbols. Here, we use a binary random variable $K_\matA$, such that $K_{\matA} = 1$, if the graph is known apriori and $0$ otherwise. Let $\mathcal{X}_{pa(i)}$ denote the set of parents of $X_{i}$.

\begin{prop}
The adjacency matrix $\matA$ is a non-zero matrix if and only if the $i$-th term in the factorization of $p(\matX|K_{\matA})$ given by $p(X_i|\mathcal{X}_{pa(i)})$ is not equal to $p(X_i)$, for any $i$.
\label{prop2}
\end{prop}
\begin{proof}
See the supplementary material.
\end{proof}

In other words, the above proposition also implies that if $\matA = 0$, then the set of parents of $X_i$ given by $\mathcal{X}_{pa(i)} = \{\}$, and hence, $p(\matX|K_{\matA}) = \prod_{i = 1}^L p(X_i)$.

\begin{prop}
If the $i$-th term in the factorization of $p(\vect,\matX|K_{\matA})$ given by $p(X_i|\mathcal{X}_{pa(i)})$ is not equal to $p(X_i)$ for any $i$, then $H(\matX) < \sum_{i = 1}^L H(X_i)$, where $H(\cdot)$ is the entropy function.
\end{prop}
\begin{proof}
See the supplementary material.
\end{proof}

From the propositions stated above, we observe that if $A(i,j) \neq 0$ for all $i,j$, then the entropy of the source is strictly smaller than entropy of the source that emits statistically independent symbols. Furthermore, if the knowledge of $\matA$ is not provided for data representation, the optimal encoder may need to consider $A(i,j) = 0$ for all $i,j$, and as a result the number of bits used to represent the source is as large as $ \sum_{i = 1}^L H(X_i)$. Therefore, it is evident that the knowledge of $\matA$ must be appropriately used for data representation of the source so that the number of bits required to encode such a source is strictly less than $ \sum_{i = 1}^L H(X_i)$. Here, we state and prove this fundamental information theoretic source encoding argument, since it provides us a strong motivation to design efficient encoders. Towards that direction, we incorporate the knowledge of $\matA$ in the context of structured survival prediction.

\subsubsection{CVAE and the Cost Function} A possible approach towards utilizing the knowledge of the adjacency matrix for source encoding is by using the variational autoencoder (VAE) \citep{Kingma2019FTML}. Several authors have successfully employed VAEs for joint source-channel coding \citep{choi2019neural}. Motivated by this, we derive a conditional variational autoencoder (CVAE) framework for DAG based survival prediction, while incorporating the knowledge of $\matA$.

\begin{figure}
    \centering
    \includegraphics[width = \textwidth]{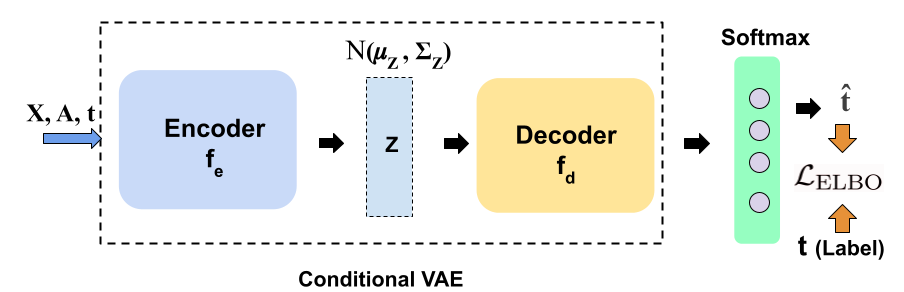}
    \caption{\texttt{DAGSurv} framework: $\matX$, $\mathbf{A}$ and $\vect$ are provided as inputs to the CVAE during training. The decoder is followed by the softmax layer, such that the output $\hat{\vect}$ represents the probability that an individual will experience an event at a given time. During test time, only the decoder($f_d$) is used where $\matX$, $\matZ$ (Input samples to decoder are from N$(0,\matI)$. The reparameterization trick ensures that $\matZ$ is sampled from N$(\mu_\matZ,\Sigma_{\matZ})$ and this distribution is embedded into decoder during training.) and $\mathbf{A}$ are provided as inputs, and $\hat{\vect}$ is obtained at the output.}
    \label{fig:DAGDeepSurvArchi}
\end{figure}

We use the standard CVAE \citep{sohn2015learning} for incorporating DAG into survival prediction. The conditional refers to the conditional probability density function (pdf) used in CVAE, instead of the joint pdf as used in VAE. Although VAE and CVAE use latent variable based formulation, conditioning on $\vecx$ is unique to CVAE. The novelty in the proposed method is in combining the knowledge of DAG and individual features for SA by encoding the DAG structure into the graph as additional information. The generative aspect of CVAE allows for the ELBO framework for encoding the graph into the neural network, and  predictive capability of DAGSurv is a result of prediction capability of CVAE. In order to design DAGSurv, we employ the sample generation process according to the  generalized SEM given by
\begin{equation}
\label{eq:sampgenlinsem}
\vect= f_{d}\left(\left(\matI-\matA^{T}\right)^{-1}g( [\matX^T,\matZ^T])\right),
\end{equation}
where $\matA^{T}$ is the transpose of matrix $\matA$, $g:\mathbb{R}^{(2L+1) \times N} \rightarrow \mathbb{R}^{(L+1) \times N}$, and $f_{d}:\mathbb{R}^{(L+1) \times N} \rightarrow \mathbb{R}^{M \times 1}$. Hence, the input to the decoder is $\matA$, and a concatanated matrix consisting of $\matX$ and $\matZ$. Here $\matZ \in \mathbb{R}^{N \times (L+1)}$ is a latent variable with a zero-mean Gaussian prior distribution N$(0,\matI)$, and $\matI$ is the identity matrix.  Often, \eqref{eq:sampgenlinsem} is referred to as the {\em decoder} model, and the corresponding {\em encoder} model is given by
\begin{equation}
\label{eq:encgenlinsem}
\matZ^T= \left(\matI-\matA^{T}\right) f_{e}(\tilde{\matX}^T),
\end{equation}
 where $f_{e}:\mathbb{R}^{(L+1) \times N} \rightarrow \mathbb{R}^{(L+1) \times N}$ is a parameterized function of the encoder and  $\tilde{\matX} \in {\re}^{N\times (L+1)} $ denotes the  augmented matrix consisting of the features in $\matX$ and time-to-event vector $\vect$, i.e., $\tilde{\matX}= [\matX, \vect]$,. Note that if $\matA = 0$ above, the encoder is given as $\matZ^T = f_e(\tilde{\matX}^T)$ and the decoder is give by $\vect = f_d(g[\matX^T ,\matZ^T])$, which is similar to the encoder and decoder correspond to the conventional CVAE, where the input covariates $\matX$ are considered statistically independent.
 
 For purposes of data-driven survival time prediction, we learn the parameters that constitute encoder and decoder by maximizing the log-evidence $\frac{1}{N}\sum_{n=1}^{N} \ln\left(p\left(t^{n}|\vecx_{n}\right)\right)$, where $\vecx_{n}$ denotes the covariates of the $n$-th sample in $\matX$. Maximizing the log-evidence is often intractable, and hence, we resort to variational inference theory which allows us to maximize the lower bound on evidence, referred to as ELBO \citep{BishopBook}. The relationship between log-evidence and ELBO is given as
\begin{align}
\frac{1}{N}\sum_{n=1}^{N} \ln\left(p\left(t^{(n)}|\vecx_{n}\right)\right) \geq \frac{1}{N}\sum_{n=1}^{N}\ex_{q\left(\vecz_{n}|\vecx_{n},t^{(n)}\right)}\left[\ln\left( \frac{p\left(t^{(n)},\vecz_{n}|{\vecx}_{n}\right)}{q\left(\vecz_{n}|{\vecx}_{n},t^{(n)}\right)}\right)\right]\label{eq:logevjen}
 \equiv \mathcal{L}_{\textnormal{ELBO}}.
\end{align}
Here, $q\left(\vecz_{n}|{\vecx}_{n},t^{(n)}\right), 1\leq n \leq N$, denotes the variational posterior distribution, which {\em encodes} the samples into the latent variable $\vecz_n$. Here, $\vecz_n$ denotes the $n$-th row of $\matZ$. Unlike the conventional VAE, the decoder in CVAE is trained to {\em predict} the target, and in this context, time-to-event $\vect$ for previously unseen samples. In particular, we obtain the mean and covariance of the conditional distribution $p(\vect|\matX,\matZ)$, and the predictions are obtained by sampling the conditional distribution.  Further, we simplify $\mathcal{L}_{\textnormal{ELBO}}$ as \citep{BishopBook}
\begin{align}
\mathcal{L}_{\textnormal{ELBO}} = \frac{1}{N}\sum_{n=1}^{N}\ex_{q\left(\vecz_{n}|{\vecx}_{n},t^{(n)}\right)}\left[\ln\left( p\left(t^{(n)}|\vecz_{n},{\vecx}_{n}\right)\right)\right]\vphantom{\ex_{q\left(\vecz_{n}|{\vecx}_{n},t^{(n)}\right)}}-D_{\textnormal{KL}}\left( {q\left(\vecz_{n}|{\vecx}_{n},t^{(n)}\right)}||{p\left(\vecz_{n}\right)}\right),\label{eq:likekl}
\end{align}
where $D_{\textnormal{KL}}(\cdot||\cdot)$ is the KL divergence function and $p(\vecz_n)$ is the prior distribution on $\vecz_n$. Hence, ELBO leads to an expected likelihood based objective function, constrained by KL-divergence. 
Since time-to-event data is censored, the 
\begin{equation}
    \ln p\left(t^{(j)}|\vecx_j, \vecz_j\right) = \delta_j \ln f\left(t^{(j)}|\vecx_j, \vecz_j\right) + (1 - \delta_j) \ln S\left(t^{(j)}|\vecx_j, \vecz_j\right),
    \label{eq:survCost}
\end{equation}
where $\delta_j$ is an indicator variable as defined earlier, $f(t|\vecx,\vecz)$ is the failure density, and $S(t|\vecx,\vecz)$ is the survival function. Here, $\hat{\vect}$ is a probability distribution $\hat{\vect} = [\hat{t}_1, \hdots, \hat{t}_{M}]$, i.e., given the covariates $\matX$, $\hat{t}_k$ is the probability that the individual will experience the event at $k$-th time-epoch, as depicted in Fig.~\ref{fig:DAGDeepSurvArchi}. Similar to \citep{lee2018deephit}, the cost function in \eqref{eq:survCost} drives the network to learn non-linear, non-proportional relationships between covariates and risks, for a given event. Hence, the overall cost function of the survival based CVAE integrates the above cost function into $\mathcal{L}_{\textnormal{ELBO}}$. 

In order to accomplish the proposed design, we use the encoder model which is a multilayered perceptron (MLP) with weights $\matW_{e}$, represented as $f_e$. Accordingly, at the decoder, $f_{d}$ is an MLP with weights $\matW_{d}$, followed by a softmax layer. The decoder of the CVAE generates the samples from the conditional distribution $p(\vect|\matZ,\matX)$, which are given by
\begin{equation}
\hat{\vect} \leftarrow \textnormal{Softmax}((I-\matA^{T})^{-1}\matZ, \matW_{d}, \matX)),
\end{equation}
where $\matZ$ is generated at encoder. The  weights $\matW_{e}$ and $\matW_{d}$, and thereby the functions $f_{e}$ and $f_{d}$  are learnt by maximizing $\mathcal{L}_{\textnormal{ELBO}}$, as given in \eqref{eq:likekl}. Since we integrate the SA based cost function given in \eqref{eq:survCost} into $\mathcal{L}_{\textnormal{ELBO}}$, it is possible to train the CVAE for efficient, graph-based, time-to-event prediction. For prediction on previously unseen samples, only the decoder is used, as shown in Fig.~\ref{fig:DAGDeepSurvArchi}.
 
 In summary, our model leads to a predictive distribution for the survival time of a user based upon the user's covariates and the underlying structure that exists among those covariates.

\section{Simulation Results}
In this section, we demonstrate the efficacy of \texttt{DAGSurv} on synthetic and publicly-available real-world clinical datasets such as METABRIC  \citep{curtis2012genomic},  Rotterdam \citep{foekens2000urokinase} $\&$ GBSG \citep{schumacher1994randomized}. These real-world datasets are a widely-accepted standard, and have been used for bench-marking several methods such as \texttt{DeepSurv} \citep{katzman2018deepsurv} and \texttt{DeepHit} \cite{lee2018deephit}. We provide the description of the datasets along with the processing steps, followed by the evaluation metric, baseline approaches and implementation specifics of \texttt{DAGSurv}. For reproducibility purposes, we have made the code public at \href{}{https://github.com/rahulk207/DAGSurv}.

\subsection{Datasets \& Data processing} 
 \subsubsection{Synthetic Datasets}
    We sample a random DAG, $\mathcal{G}(V,E)$ using Erdos-R\'{e}nyi model \citep{erdds1959random}, where, $|V| = L+1$, $L$ refers to the number of covariates and $1$ refers to the target variable which is the time-to-event outcome. For simulations, we set the expected node degree as $3$. We initialise the edge weights uniformly but randomly, i.e., as $\forall e\in E$, we have the DAG edge weight $\mathcal{W}(e) \sim \textnormal{U}(0.5, 2)$. We embed the DAG-based  relationship among covariates using the following equations \citep{YuCGYicml19}:
 \begin{align}
    \matX^T = \matA^T(\cos (\tilde{\matX}+\mathbf{1})) + \matZ_\matX^T, \quad \textnormal{and} \quad \vect = \max(\mathbf{0}, c\exp\left\{{\matA^T(\cos (\tilde{\matX}^T+\mathbf{1}))}\right\} + \matZ_\vect^T),
 \end{align}
where entries of $\matZ_\matX$ and $\matZ_\vect$ are sampled independently from N$(0,1)$ and N$(30,70)$, respectively. Further, $\mathbf{1}$ is an all $1$ matrix, $\mathbf{0}$ is an all zero matrix, and $c$ is a constant chosen such that we obtain $\vect$ in a certain range;  we set $c = 90$. Using this data generating process, we obtain $10,000$ data points. Although harsh and conservative, we censored $50\%$ of the data, and we sample censoring time uniformly but randomly as $\textnormal{U}(0, \max(t))$. Using the above described settings, we created the following two datasets - 
\begin{enumerate}
    \item \textbf{Synthetic-small}: Here, we set $L = 9$ covariates (hence, $|V| = 10$).  
    \item \textbf{Synthetic-large}: In order to test our model's scalability and performance on a bigger dataset, we set $L = 49$.
\end{enumerate}
\subsubsection{Real-world Datasets} 
\begin{figure}[ht]
    \centering
    \begin{minipage}{.5\textwidth}
        \centering
        \includegraphics[scale = 0.30]{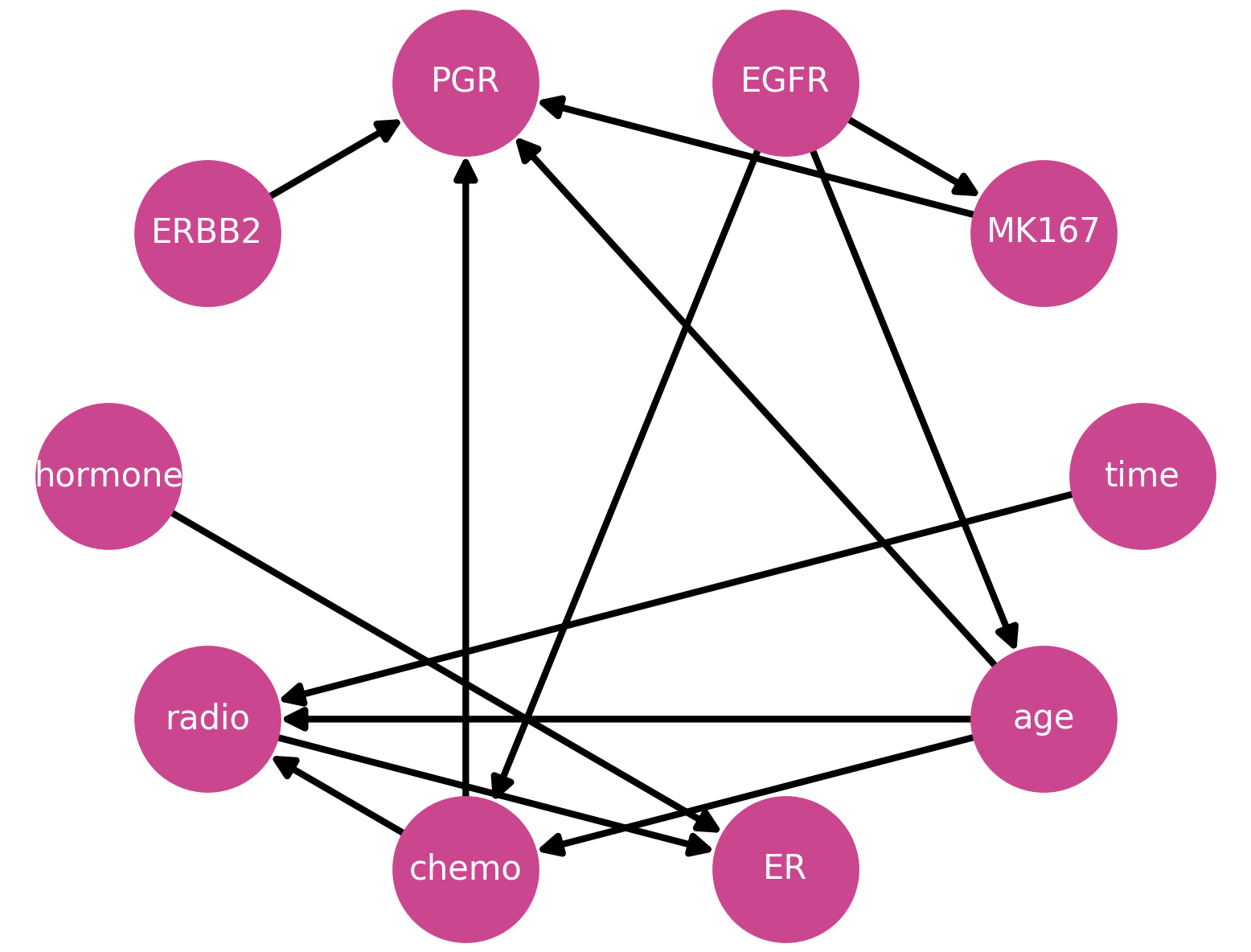}
        \caption{DAG: METABRIC}
        \label{fig:DAG_Metabric}
    \end{minipage}%
    \begin{minipage}{0.5\textwidth}
        \centering
        \includegraphics[scale = 0.30]{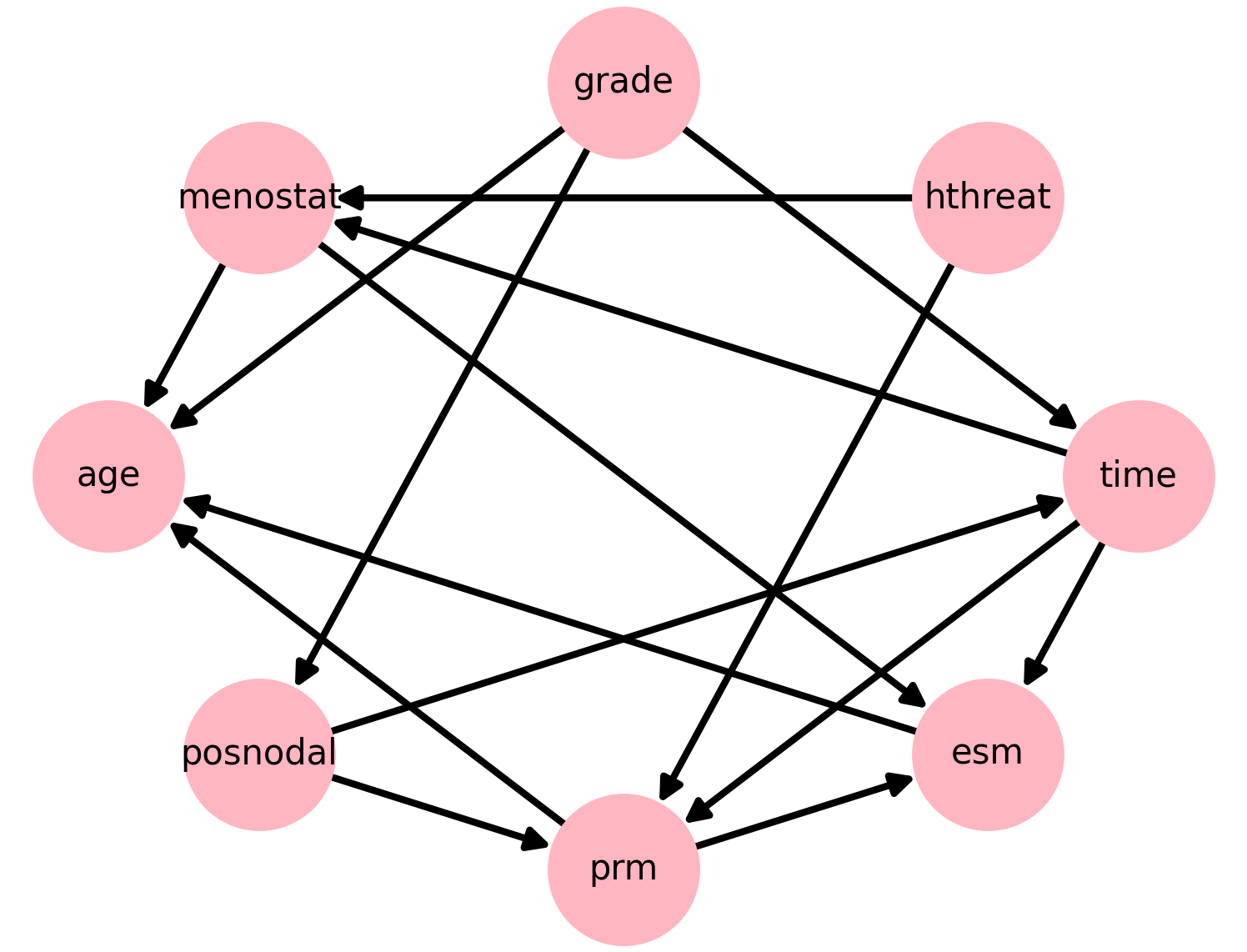}
        \caption{DAG:GBSG}
        \label{fig:DAG:GBSG}
    \end{minipage}
\end{figure}

\begin{table*}[t]
    \centering
    \begin{tabular}{|p{3cm}|p{2cm}|p{2cm}|p{2cm}|p{1.5cm}|p{1.5 cm}|}
    \hline
     Dataset & $\#$ Censored & $\#$ Features & $T_{max}$ & $C_{max}$ \\
    \hline
    \hline
    \texttt{Synthetic-small}  & 50.06\% & 9 & 377 & 91\\
    \hline
    \texttt{Synthetic-large}  & 51.58\% & 49 & 395 & 235\\
    \hline
    \texttt{METABRIC}  & 42.06\% & 9 & 355 & 337\\
    \hline
    \texttt{GBSG}  & 43.23\% & 7 & 83 & 87\\
    \hline
\end{tabular}
\caption{Description of Synthetic and Real-world Datasets ($C_{max}$ is the maximum Censoring Time).}
\label{tab:Datasets}
\end{table*}
\begin{table*}[t]
    \centering
    \begin{tabular}{|c|c|c|c|c|c|}
    \hline
     Dataset & $n_{l}$,$n_{h}$(Encoder) & $n_{l}$,$n_{h}$(Decoder) & Activation & lr \\
    \hline
    \hline
    \texttt{Synthetic-small} & 5,128 & 3,64 & ReLU & 1e-4\\
    \texttt{Synthetic-large} & 5,64 & 4,32 & ReLU & 1e-5\\
    \texttt{METABRIC} & 3,256 & 3,64 & SELU & 1e-5\\
    \texttt{GBSG} & 3,128 & 3,32 & ReLU & 1e-5\\
    \hline
\end{tabular}

\caption{Hyperparameters used in different datasets: $n_l$ and $n_h$ represent the number of layers and number of hidden nodes per layer, respectively and lr is the learning rate.}
\label{tab:Hyperparams}
\end{table*}

In the context of real-world datasets, the input DAG is not known. Given the covariates in a dataset, manually constructing a DAG may be infeasible since it requires domain-specific expertise, and hence, it can be an expensive process. Instead, we used two well-known algorithms for pre-computing our adjacency matrix $\matA$. They are as follows:
\begin{enumerate}
    \item \textbf{bnlearn, R-package} \citep{scutari2009learning} - Within the R package, we used the Hill Climbing (HC) algorithm to learn the structure of Bayesian network, which leads to an unweighted directed graph.
    \item \textbf{DAG-GNN} \citep{YuCGYicml19} - DAG-GNN is a recent deep-learning model for generating a weighted DAG, establishing structure among the features of a given dataset.
\end{enumerate}
We use these algorithms on the real-world datasets as follows: 
\begin{itemize}
    \item \textbf{METABRIC:} The Molecular Taxonomy of Breast Cancer International Consortium (METABRIC) is a clinical dataset which consists of gene expressions used to determine different subgroups of breast cancer. We consider the data for 1,904 patients with each patient having 9 covariates - $4$ gene indicators (MKI67, EGFR, PGR, and ERBB2) and $5$ clinical features (hormone treatment indicator, radiotherapy indicator, chemotherapy indicator, ER-positive indicator, age at diagnosis). Furthermore, out of the total 1,904 patients, 801 $(42.06 \%)$ are right-censored, and the rest are deceased (event). We obtained the DAG as depicted in Fig.~\ref{fig:DAG_Metabric} using a modified DAG-GNN algorithm. 
    \item \textbf{GBSG:} Rotterdam and German Breast Cancer Study Group (GBSG) contains breast-cancer data from Rotterdam Tumor bank. The dataset consists of 2,232 patients out of which 965 $(43.23 \%)$ are right-censored, remaining are deceased (event), and there were no missing values. In total, there were 7 features per patient that include hormonal therapy (hthreat), age of patient, menopausal status, tumor grade, number of positive nodes, progesterone receptor(in fmol) and estrogen receptor(in fmol). The graph for this dataset is obtained using bnlearn and it is as depicted in Fig.~\ref{fig:DAG:GBSG}. 
\end{itemize}

\subsection{Implementation and Evaluation} In this section, we provide the details of the experimental evaluation, which includes the evaluation metric, baseline models, implementation specifics and the experimental results. We randomly split the data into training set ($80\%$) and test set ($20\%$), and further reserved $20\%$ of the training set for validation. 

As depicted in Fig.~\ref{fig:DAGDeepSurvArchi}, \texttt{DAGSurv} is a CVAE consisting of MLPs as encoder and decoder. The model has a DNN architecture, and we used grid-search to perform an extensive hyperparameter search on the number of layers, number of hidden units, activation function and learning rate. The hyperparameter values that were used to obtain the results reported in this paper are as given in Table~\ref{tab:Hyperparams}. Adaptive Moment Estimation (Adam) was chosen as the gradient descent optimization algorithm, and the entire module was coded using pyTorch. Following the implementation in DAG-GNN \citep{YuCGYicml19}, we set the variance of the latent variable $\Sigma_{\matZ}$ as $\mathbf{I}_{L+1}$ which is the identity matrix in $L+1$ dimensions. We have considered only $\mu_{\matZ}$ as trainable, since it was observed that the value of $\Sigma_{\matZ}$ explodes due to the exponent term, particularly in datasets with larger time-to-event values. Note that the results remain unaffected in spite of this modification. 

\subsubsection{Evaluation Metric}
We employ the time-dependent concordance index (CI) as our evaluation metric since it is robust to changes in the survival risk over time. Mathematically it is given as
\begin{align}
  C_{td} &= P\left(F(t^{(i)}|x^{(i)})>F(t^{(i)}|x^{(j)})|t^{(i)} < t^{(j)}\right) \approx \frac{\sum_{i \neq j} R_{i,j}\mathbf{1}\left(F(t^{(i)}|x^{(i)})>F(t^{(i)}|x^{(j)})\right)}{\sum_{i \neq j} R_{i,j}},
\end{align}
where $\mathbf{1}\left(.\right)$ is the indicator function and  $R_{i,j} \triangleq\mathbb{1}\left(t^{(i)} < t^{(j)}\right)$, i.e., we use an empirical estimate of the time-dependent CI as our metric \citep{lee2018deephit}. To test the robustness of trained models on unseen data, we performed bootstrapping on the test set. Using the bootstrap $C_{td}$ values obtained on the test set, we plot notched box-plots and compared them for the proposed and baseline methods. The notch here represents $95\%$ confidence interval ($C_I$) around the median which can be calculated as $\textnormal{median} \pm 1.57\times \frac{IQR}{\sqrt{b}}$, where IQR is the interquartile range which includes $50\%$ of the data and $b$ denotes the number of bootstrap samples. 
  
\subsubsection{Baseline Models}
In this section, we discuss the following baseline approaches for survival prediction against which we compare the proposed \texttt{DAGSurv}:
\begin{itemize}
    \item \textbf{\texttt{CoxTime}:} Cox-PH is a classical, and one of the most fundamental baselines to compare against. While the PH assumption is essential for these models, they allow easy interpretation and ranking of risk factors. We used \texttt{CoxTime} \citep{kvamme2019time} which is a relative risk neural network model that extends Cox regression beyond linear PH. 
    \item \textbf{\texttt{DeepSurv}:} \texttt{DeepSurv} is a DNN extension of the classical Cox-PH model. It generally performs better than Cox-PH model since it captures some non-linearity which may be important in the context of real-world datasets. 
    \item \textbf{\texttt{DeepHit}:} \texttt{Deephit} predicts the time-to-event directly, unlike survival risk prediction algorithms such as \texttt{DeepSurv}/\texttt{Cox}. Furthermore, \texttt{Deephit} is not inherently based on the PH assumption, and hence, an important baseline to compare against. 
\end{itemize}

\subsection{Experimental Results}
In this section, we illustrate the time-dependent CI values ($C_{td}$), along with the confidence intervals ($95\%$) using tables and box-plots. 

\subsubsection{Synthetic Dataset}
In this section, we present the results obtained using the proposed and baseline methods on a small and large synthetic datasets which we defined in the previous section. It is observed that most of the models find it hard to learn the underlying model, and the $C_{td}$ values as measured on the test-set are low. It can be observed from Table~\ref{tab:simulateddata} that \texttt{Deepsurv} and \texttt{CoxTime} fail to learn a meaningful model and their $C_{td}$ values are close to $0.5$. With fewer model-based assumptions, \texttt{DeepHit} and \texttt{DAGSurv} are able to learn the model with acceptable $C_{td}$. Note that we do not employ the constraint on concordance index as in Deephit. Generally this constraint is hard to compute for large datasets, since it requires pairwise computations. The knowledge of the input DAG helps \texttt{DAGSurv} to perform better than \texttt{DeepHit}, in the absence of the concordance constraint. As expected, the box-plot shows smaller variation in values of $C_{td}$ over the test set since in the case of synthetic data, the testing and training samples come from the same data generating process. 

\begin{table}[t]
\begin{center}
\begin{tabular}{|c|c|c|c|}
    \hline
    \multicolumn{2}{|c|}{Synthetic-small}&\multicolumn{2}{|c|}{Synthetic-large}\\
    \hline
    Algorithms & $C_{td}$ ($95\%~~C_I$) & Algorithms & $C_{td}$ ($95\%~~C_I$) \\
    \hline
    \hline
    \texttt{DAGSurv} & $\mathbf{0.5692} \pm 0.0009$ & \texttt{DAGSurv} & $\mathbf{0.5396} \pm 0.0006$\\
    \texttt{DeepHit} & $0.5532 \pm 0.0009$ & \texttt{DeepHit} &  $0.5326 \pm 0.0005$ \\
    \texttt{DeepSurv} & $0.4956 \pm  0.0005$ & \texttt{DeepSurv} & $0.4936 \pm 0.0004$\\
    \texttt{CoxTime} & $0.5134 \pm  0.0005$ &\texttt{CoxTime} & $0.5045 \pm 0.0005$\\
    \hline
\end{tabular}
\end{center}
\caption{$C_{td}$ for Synthetic-small and Synthetic-large datasets}
\label{tab:simulateddata}
\end{table}

\begin{figure}[!htb]
    \centering
    \begin{minipage}{.5\textwidth}
        \centering
        \includegraphics[scale = 0.30]{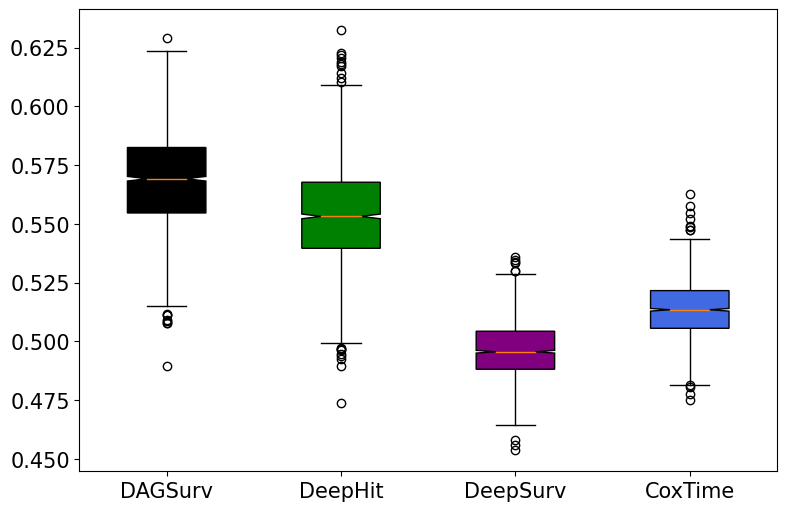}
        \caption{Box-plot: $C_{td}$ for Synthetic-small}
        \label{fig:boxplot_synsmall}
    \end{minipage}%
    \begin{minipage}{0.5\textwidth}
        \centering
        \includegraphics[scale = 0.30]{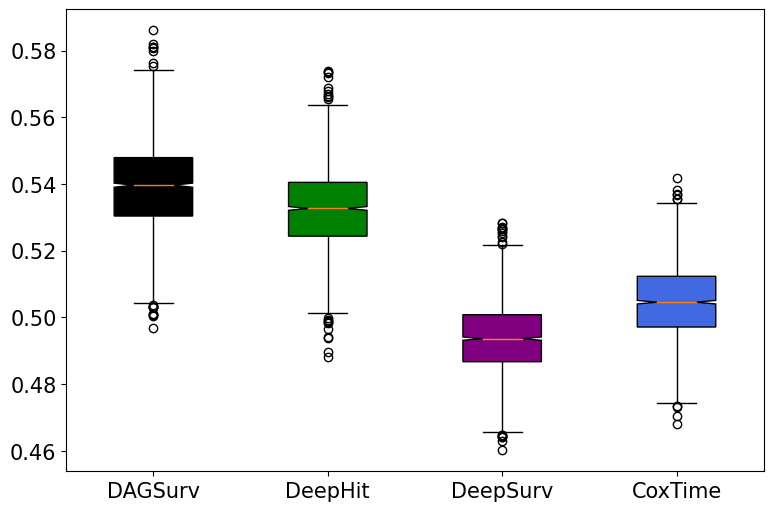}
        \caption{Box-plot: $C_{td}$ for Synthetic-large}
        \label{fig:boxplot_synlarge}
    \end{minipage}
\end{figure}

\subsubsection{Real-world datasets}
In this section, we illustrate the performance of the proposed approach and the baseline schemes on real-world datasets which we described earlier. We observe that \texttt{DAGSurv} consistently performs better or is as competitive as compared to the baseline schemes. 

In addition to improved performance, \texttt{DAGSurv} lends itself to better interpretation as well. First of all, the concordance score acts as validation for the input graph, i.e., if $C_{td}$ improves when we set $\matA = 0$ in \texttt{DAGSurv}, it implies that graph is not aiding to obtain better ML models for survival analysis. Further, it also helps to establish relationship between covariates and the outcome. For instance, we observe from the graph pertaining to the GBSG dataset in  Fig.~\ref{fig:DAG:GBSG} that the grade of tumor affects both, the number of positive lymph nodes as well as time-to-event (death). Hence, the relationship between number of positive lymph nodes and survival time, would have to account for the grade of tumor. Such interpretable results are powerful tools for practitioners and clinicians, and we plan to explore the aspects of explainable AI in our future work.

\begin{table}[t]
\begin{center}
\begin{tabular}{|c|c|c|c|}
    \hline
    \multicolumn{2}{|c|}{METABRIC}&\multicolumn{2}{|c|}{GBSG}\\
    \hline
    Algorithms & $C_{td}$ ($95\%~~C_I$) & Algorithms & $C_{td}$ ($95\%~~C_I$) \\
    \hline
    \hline
    \texttt{DAGSurv} & $\mathbf{0.7323} \pm 0.0056$ & \texttt{DAGSurv} & $\mathbf{0.6892} \pm 0.0023$\\
    \texttt{DeepHit} & $0.7309 \pm 0.0047$ & \texttt{DeepHit} &  $0.6602 \pm 0.0026$ \\
    \texttt{DeepSurv} & $0.6575 \pm 0.0021$ & \texttt{DeepSurv} & $0.6651 \pm 0.0020$\\
    \texttt{CoxTime} & $0.6679 \pm 0.0020$ &\texttt{CoxTime} & $0.6687 \pm 0.0019$\\
    \hline
\end{tabular}
\end{center}
\caption{$C_{td}$ for METABRIC and GBSG datasets}
\label{tab:multicol}
\end{table}

\begin{figure}[!htb]
    \centering
    \begin{minipage}{.5\textwidth}
        \centering
        \includegraphics[scale = 0.30]{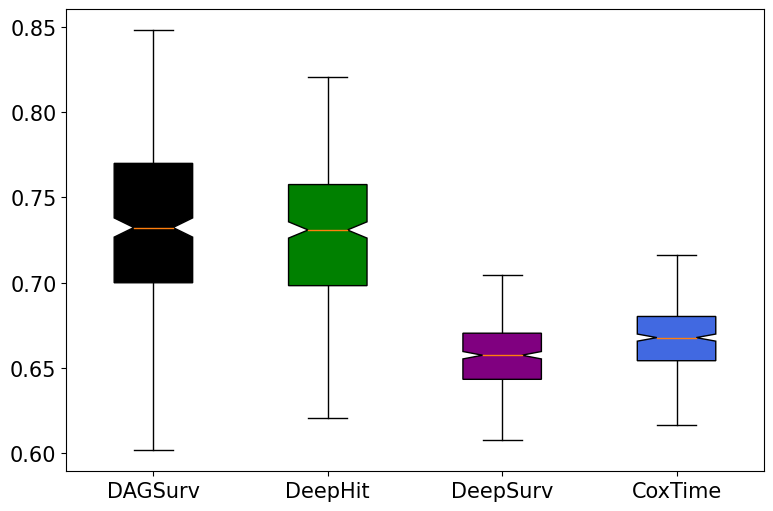}
        \caption{Box-plot: $C_{td}$ for METABRIC}
        \label{fig:boxplot_Metabric}
    \end{minipage}%
    \begin{minipage}{0.5\textwidth}
        \centering
        \includegraphics[scale = 0.30]{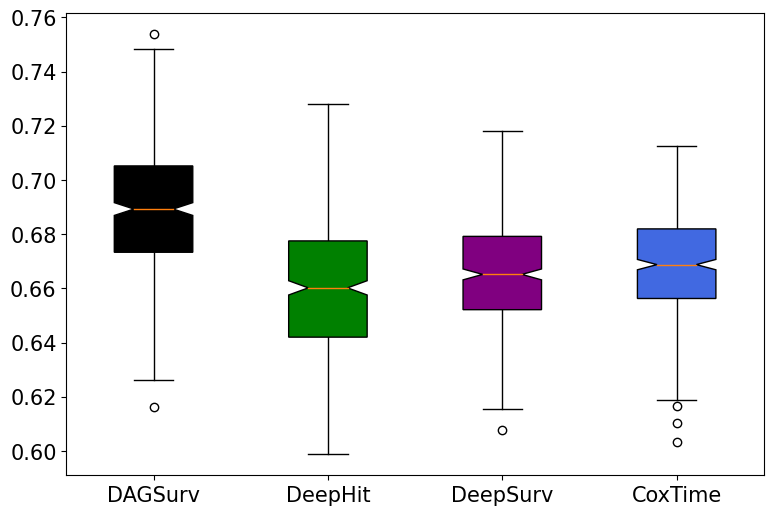}
        \caption{Box-plot: $C_{td}$ for GBSG}
        \label{fig:boxplot_GBSG}
    \end{minipage}
\end{figure}

\subsection{Discussions and Conclusions}

In this work, we propose \texttt{DAGSurv}, in which we incorporate the knowledge of the causal DAG and design a novel CVAE framework for SA. Using the source coding argument we prove that the knowledge of the DAG leads to reduced entropy as compared to a source that emits statistically independent symbols, a default choice in DAG-agnostic ML models. We employed the CVAE as a possible source encoder for achieving efficient data representation. However, CVAE is not an optimal choice, and we reserve the the design of optimal source encoder to future work. 

Using synthetic and real-world datasets, we demonstrated that \texttt{DAGSurv} has an improved performance (in terms of concordance index) while it being more interpretable. Using this method requires the knowledge of the DAG, which is generally not known. In the absence of experts' knowledge, we demonstrated that one may opt to use one of the several algorithms available to obtain a DAG from a given dataset. Unlike \texttt{CoxTime} and \texttt{DeepSurv}, \texttt{DAGSurv} can be used in the presence of time-varying hazard. Further, note that \texttt{DAGSurv} does not require the expensive concordance index based constraint which requires pairwise comparisons across all the points in a dataset as in \citep{lee2018deephit}. In spite of not using the constraint, \texttt{DAGSurv} is  able to perform better than \texttt{DeepHit}. Furthermore, \texttt{DAGSurv} can be used to validate the causal relations in any graphical model. 

Further, extending our analysis to the multiple risk case is straightforward. Some interesting extensions include analysis in the context of recurring events \citep{CRESA} and for counterfactual inference. 

\bibliography{acml21}

\end{document}